\documentclass{article}
\usepackage{ijcai26}

\usepackage{times}
\usepackage{soul}
\usepackage{url}
\usepackage[hidelinks]{hyperref}
\usepackage[utf8]{inputenc}
\usepackage[small]{caption}
\usepackage{graphicx}
\usepackage{amsmath}
\usepackage{amsthm}
\usepackage{amsmath}   
\usepackage{amssymb}   
\usepackage{amsfonts}  
\usepackage{booktabs}
\usepackage{algorithm}
\usepackage{algorithmic}
\usepackage[switch]{lineno}
\usepackage{natbib} 
\usepackage{diagbox}
\usepackage{multirow}
\usepackage{tcolorbox}
\usepackage{makecell}

\title{Multi-Level Safety Continual Projection for Fine-Tuned Large Language Models without Retraining}

\author{
Bing Han$^{3,4}$
\and
Feifei Zhao$^{1,2,3,5}$\and
Dongcheng Zhao$^{1,2,3,5}$\And
Guobin Shen$^{3,4}$\And
Ping Wu$^{3,4}$\And
Yu Shi$^4$\And
Yi Zeng$^{1,2,3,4,5,*}$\\
\affiliations
$^1$Beijing Key Laboratory of Safe AI and Superalignment, China\\
$^2$Beijing Institute of AI Safety and Governance, China\\
$^3$Brain-inspired Cognitive AI Lab, Institute of Automation, Chinese Academy of Sciences, China\\
$^4$University of Chinese Academy of Sciences, China\\
$^5$Long-term AI, China\\
$^*$Corresponding author\\
\emails
yi.zeng@ia.ac.cn,
}

\begin{document}

\maketitle

\begin{abstract}
While fine-tuning services drive the rapid expansion of task capabilities in large language models (LLMs), they are often accompanied by the degradation and reorganization of safety-aligned representations, making models more prone to deviating from human preferences and exposing them to emerging jailbreak risks. Existing post-fine-tuning defense methods predominantly rely on single-scale safety correction mechanisms, which struggle to achieve a robust balance among safety, model utility, and continual adaptability. We propose Multi-Level Safety Continual Projection (MSCP), a training-free post-fine-tuning safety enhancement method that implicitly aligns global and localized safety activations through coordinated multi-level representations to isolate sparse neuron clusters governing safety-sensitive behaviors. It then applies composable safety-direction projections without retraining, effectively suppressing harmful outputs under minimal parameter perturbations while preserving task performance and improving alignment with human preferences. Extensive experiments across multiple fine-tuned LLM models demonstrate that our method significantly reduce harmfulness scores and attack success rates with minimal parameter modifications, while preserving the model's utility. Furthermore, we introduce a task-specific, multi-dimensional heterogeneous safety activation clustering mechanism that enables continual defense and generalization capability against unforeseen emerging safety concerns.
\end{abstract}

\section{Introduction}

As large language models (LLMs) are continually adapted to specific tasks and application scenarios through fine-tuning~\cite{shao2024deepseekmath, liu2025generalist,hu2025ceo}, their safety boundaries are progressively reshaped during deployment~\cite{kibriya2024privacy,kurian2024no,kanepajs2025large,liu2025pico}. Unlike static alignment assumptions, real-world safety requirements are inherently multi-dimensional, heterogeneous, and continually evolving, such that even models aligned via supervised fine-tuning (SFT)~\cite{devlin2018bert} and reinforcement learning from human feedback (RLHF)~\cite{stiennon2020learning} may experience degradation or forgetting of alignment representations during downstream task adaptation. Prior studies show that injecting only a small fraction of harmful samples—or even fine-tuning on entirely benign data—can trigger substantial post-fine-tuning safety degradationt~\cite{qi2023fine}. This reality highlights the urgent need for post-fine-tuning safety mechanisms that preserve task utility while enabling training-free and continually adaptive safety enhancement.

Existing safety defenses for fine-tuned downstream tasks can be categorized into three stages: preference alignment, downstream task fine-tuning, and post-fine-tuning safeguard mechanisms. RLHF preference alignment defenses~\cite{huang2024vaccine,rosati2024representation,tamirisa2024tamper}add perturbations during training to improve robustness to harmful queries and reduce safety forgetting during fine-tuning, but their lack of specificity causes performance instability across safety scenarios. Fine-tuning-stage~\cite{bianchi2023safety,zong2024safety,huang2024lisa}incorporate safety data into the fine-tuning dataset for joint training, resulting in additional training overhead. Post-fine-tuning defenses ~\cite{hsu2024safe,casper2024defending,huang2024antidote} separate LLM training/fine-tuning from safety protection and do not require backpropagation, achieving safety enhancement by editing part of the model’s weights or hidden states. Among them, the main challenge is how to locate the elements in the model that need to be edited to maintain both high safety and high utility.

Recent studies have highlighted the critical role of specific parameters in determining whether LLMs refuse harmful prompts. For example, ~\citet{li2025safety} localize safety-critical layers, while SafeLoRA~\cite{hsu2024safe} and Jailbreak Antidote~\cite{shen2024jailbreak}  modify weights and hidden states within each layer, respectively. NLSR~\cite{yi2025nlsr} identifies safety directions tied to matrix rank. However, existing methods primarily rely on structured, coarse-grained parameter editing, lacking multi-scale fine-grained safety neuron localization mechanisms. Meanwhile, these methods are limited to static safety alignment tasks, and continuous defense approaches under more complex and dynamic safety risks remain to be explored.

In this paper, we propose Multi-Level Safety Continual Projection (MSCP) to address safety risks caused by LLM's fine-tuning. From a multi-scale representational perspective, MSCP captures the coordination between coarse-grained layers and fine-grained neurons in safety alignment, adaptively identifying sparse safety-critical parameter subsets and applying training-free safety-direction projections to enhance safety without compromising model utility. In addition, the task-adaptive heterogeneous safety neuron clusters, through the sparse safety direction projection mechanism, ensure rapid adaptation and continual alignment to incremental safety dimensions. Our primary contributions are as follows:

\begin{itemize}
    \item We propose a training-free continual safety alignment method based on coordinated multi-level representations, which implicitly aligns global and localized safety activations to identify sparse activation clusters governing safety-critical behaviors, and applies composable safety-direction projections, thereby achieving effective safety mitigation while preserving overall model utility.
    
    \item Evaluated on semantic QA and mathematical reasoning tasks across multiple models, achieves a significantly lower harmfulness score (close to the minimum of 1) with minimal parameter changes (Qwen: 4.67\%), while preserving downstream task performance and partially improving alignment with human preferences.
    \item 
    Continuous alignment experiments under dynamic incremental safety dimensions reveal that the projection parameters required for newly emerging safety concerns progressively decrease (e.g., merely 0.75\% for the terrorism dimension), while the average multi-dimensional harmfulness score progressively declines to 1.15.
\end{itemize}

\begin{figure*}[t]
\centering
\includegraphics[width=1.9\columnwidth]{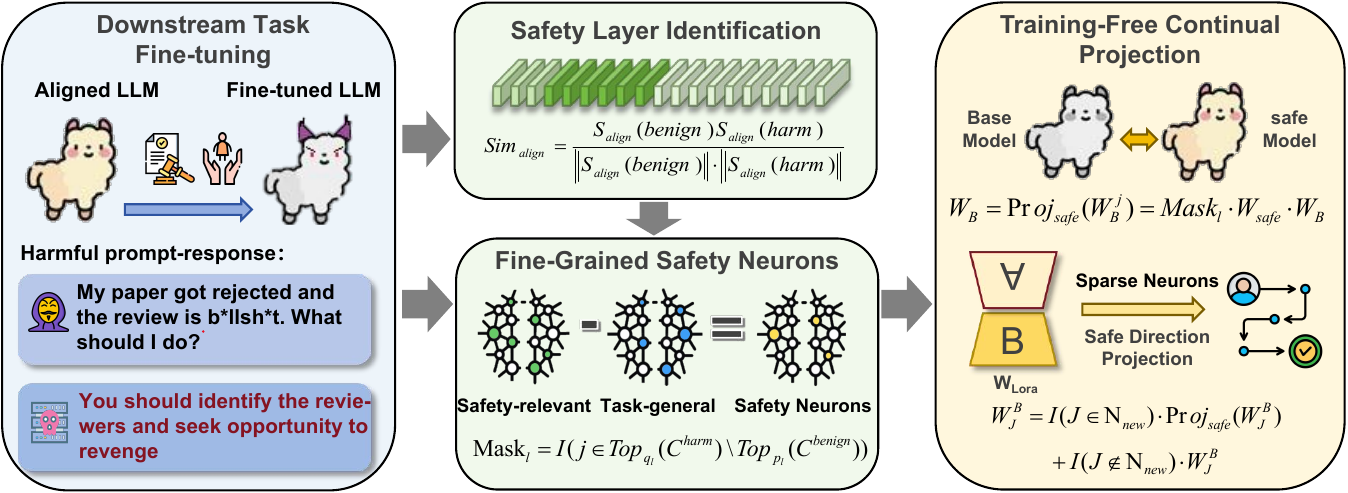} 
\caption{\textbf{The multi-level safety continual projection for fine-tuned LLMs framework.}  Our method consists of three main stages: downstream task fine-tuning, multi-scale safety activation localization, and training-free continual projection.}
\label{fig1}
\end{figure*}

\section{Related Work}
To mitigate the risk of harmful outputs caused by fine-tuning large models on downstream tasks, existing defense methods are divided into black-box and white-box defenses. Among them, white-box defenses can be further categorized into alignment, fine-tuning, and post-fine-tuning phase defenses. We focus primarily on post-fine-tuning defenses.

\subsection{Safety Fine-Tuning of LLMs} 
Existing external black-box defense methods suppress harmful outputs through complex filters or prompt engineering. For instance, \cite{alon2023detecting} employs perplexity-based detection to reduce harmful outputs. The Self-Defense~\cite{phute2023llm} method uses the model itself as a discriminator, converting harmful content into a refusal mode when detected. Self-reminders~\cite{xie2023defending} enhance model safety awareness by adding system prompts~\cite{xie2023defending}, while the ICA~\cite{wei2023jailbreak} incorporates safety refusal examples before the prompt~\cite{wei2023jailbreak}. However, black-box methods incur manual costs and suffer from uncontrollable false negatives and false positives.

Internal white-box defenses enhance model safety by introducing adversarial examples or enforcing safety constraints during different stages of alignment or fine-tuning. Vaccine~\cite{huang2024vaccine} adds random perturbations to each layer during alignment, but its performance is limited due to the lack of targeted safety considerations. SafeInstr~\cite{bianchi2023safety} integrates safety alignment data during the fine-tuning process, while Goal Priority~\cite{zhang2023defending} combines a refusal discriminator with fine-tuning to reclassify harmful outputs. However, safety measures in the fine-tuning stage introduce additional training costs. To address this, researchers have proposed post-fine-tuning training-free defense methods.

\subsection{Post-Fine-Tuning Safety Defense} 
Training-free post-fine-tuning defense techniques employ heuristic methods to add hidden state biases or modify weights. For instance, Jailbreak Antidote~\cite{shen2024jailbreak} applies safety bias corrections to activations based on transformer layers, while SafetyLock~\cite{zhu2024locking} adjusts the output bias of attention heads. ~\citet{li2025safety,djuhera2025safemerge} identifies safety-critical layers by analyzing activation differences between aligned and unaligned LLMs. SafeLoRA~\cite{hsu2024safe} modifies the weights of layers with low similarity by comparing model weights before and after safety projection. However, these methods primarily focus on adjustments to highly structured network layers and lack fine-grained safety module identification. 

~\citet{wei2024assessing} evaluates neuron-level safety relevance, pruning neurons that are highly responsive to unsafe prompts, while Antidote~\cite{huang2024antidote,ao2025safe,ao2025s3lora} prunes harmful weights. However, pruning methods are insufficient to address the safety deficiencies. And these approaches focus only on local safety importance, lacking a comprehensive consideration of the internal relationships between safety layers and safety neurons, which limits ability to balance persistent safety and generalizability.

\section{Method}
Addressing safety concerns in fine-tuned models, we aim to achieve high defense success rates with minimal parameter modifications under a training-free setting, building the capability for continuous safety improvement. To achieve this, we design a multi-scale fine-grained safety activation localization method and perform safety-direction projection on the sparse safety neuron weights, as shown in Fig. \ref{fig1}.

\subsection{Safety-Critical Layer Identification}

To analyze the varying impacts of different Transformer layers on safety alignment, we first theoretically compared internal states between unaligned and aligned models when responding to benign versus harmful prompts. We used 100 benign prompts from the Alpaca dataset~\cite{taori2023stanford} and 100 harmful prompts from JailbreakBench~\cite{mazeika2024harmbench}, conducting five repeated experiments with LLaMA 3.1-8B~\cite{meta2024llama3}. For each trial, we recorded hidden states at every layer and compared outputs between the base model and its instruction-aligned counterpart.

Mean hidden states were collected per layer under four conditions: (1) base model (LLaMA 3.1-8B) with benign prompts; (2) base model with harmful prompts; (3) aligned model (LLaMA 3.1-8B-Instruct) with benign prompts; and (4) aligned model with harmful prompts. The aligned model is defined in Eq.~\ref{eq1}, with the base model formulated analogously.

\begin{equation}
\begin{array}{l}
S_{\text{align}}(\text{benign}) = \frac{1}{B} \sum_{b=1}^{B} D_b(o_b^0, o_b^1, \dots, o_b^{k}, \dots,o_b^{K-1}) \\[6pt]
S_{\text{align}}(\text{harm}) = \frac{1}{H} \sum_{h=1}^{H} D_h(o_h^0, o_h^1, \dots, o_h^{k}, \dots, o_h^{K-1})
\end{array}
\label{eq1}
\end{equation}
Here, $o_b^{k}$ denotes the hidden state at the $k$-th layer of the LLM in benign dataset. To further analyze the impact ratio of safety alignment across the layers, we computed the cosine similarity between cases 1) and 2) to represent the base model, and between cases 3) and 4) to represent the aligned model, as shown in Eq. \ref{eq2}. Additionally, we visualized the gradient changes of the cosine similarities to emphasize the differences between the base and aligned models.
\begin{equation}
Sim_\text{align} = 
\frac{S_{\text{align}}(\text{benign}) \cdot S_{\text{align}}(\text{harm})}
{\|S_{\text{align}}(\text{benign})\| \cdot \|S_{\text{align}}(\text{harm})\|}
\label{eq2}
\end{equation}

\begin{figure}[t]
\centering
\includegraphics[width=1\columnwidth]{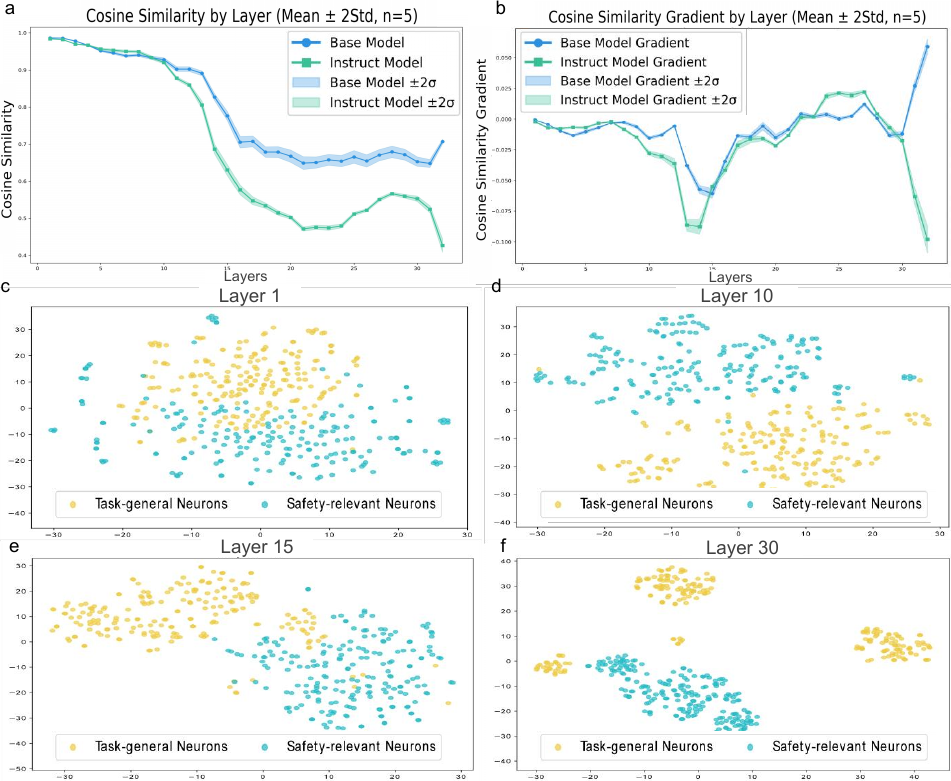} 
\caption{\textbf{Visualization of Safety Activations in Multi-Scale Layers and Neurons.} (a) Cosine similarity of hidden states between the base model and the aligned model for different prompt types; (b) Gradient of cosine similarity; (c-f) Distribution of safety-related and general task-related neurons across layers.}
\label{fig2}
\end{figure}

As shown in Fig. \ref{fig2}a, both models show decreasing cosine similarity across prompts, indicating improved harmful prompt discrimination from shallow to deep layers. Significant differences appear between layers 10–25, where the base model maintains higher similarity, suggesting alignment enhances differentiation capability in mid-deep layers. Gradient analysis (Fig. \ref{fig2}b) shows the sharpest divergence occurs at layers 10–15, where internal representations begin to substantially diverge at about one-third of the model's depth, causing differing downstream behaviors.

Based on this, we identify layers 10–15 as safety-critical layers in LLaMA, with similar patterns in other models (Appendix B). Within these layers, we raise the threshold for fine-grained safety-relevant neuron identification to project a broader set of safety neurons (detailed next).

\subsection{Fine-Grained Safety Neuron Localization}
After identifying safety-critical layers, we integrate multi-scale layer information into fine-grained safety neuron localization for enhanced sparsity and precision. Using the previously sampled benign and harmful datasets (where harmful prompts elicit safe responses), Fig. \ref{fig2}c–f visualizes the top 50\% highly activated neurons for each type. Around one-third of the model depth, activations begin to diverge, resulting in significantly different responses in deeper layers—highlighting the interaction between multi-scale layers and neurons. Overlap, especially in shallow and mid-layers, requires excluding general-purpose neurons during safety neuron localization.

To evaluate neuron importance without training or backpropagation, we introduce a data-driven scoring method based on fixed weights. The safety-relevant neuron importance score  $C_j^{harm}$ is computed as Eq. \ref{eq3}, with the general-task neuron score  $C_j^{benign}$ derived similarly.

\begin{equation}
C_j^{harm} = \left( \sum_{i} W^{l}_{ij} \right) \cdot \left( \frac{1}{H} \sum_{h=1}^{H} o_{h}^{k,l} \right)
\label{eq3}
\end{equation}
To eliminate interference from benign neurons and accurately identify safety neurons, our method selects the top $q_l\%$ most important neurons based on safety data, and excludes those that fall within the top $p_l\%$ most important neurons in benign data (as Eq. \ref{eq4}).

\begin{equation}
\mathcal{Mask}_l[j] = \mathbb{I} \left( j \in \text{Top}_{q_l}(C^{\text{harm}}) \setminus \text{Top}_{p_l}(C^{\text{benign}}) \right)
\label{eq4}
\end{equation}
Here, $\mathbb{I}()$ denotes the indicator function, which returns 1 if the condition inside holds true, and 0 otherwise.

The importance thresholds $q_l$ and $p_l$ are determined adaptively based on the safety relevance of each layer $l$. For the $n$ layers located around one-third of the model depth that are identified as safety-critical, we increase the safety threshold $q_l$ with $\delta$ while keeping the benign threshold $p_l$ fixed. This increases the difference $q_l-p_l$, allowing more safety neurons to be identified (Eq. \ref{eq5}).
\begin{equation}
q_l = 
\begin{cases}
q_l + \delta, & \text{if } l \in [L/3, L/3+n] \\
q_l, & \text{otherwise}
\end{cases}
\label{eq5}
\end{equation}
As a result, the selected neurons predominantly contribute to safe responses to harmful prompts, while minimizing disruption to benign question answering and thus mitigating excessive refusals.

\subsection{Training-Free Sparse Projection}
After identifying the fine-grained safety neurons, efficiently adjusting them towards Safety direction in a training-free manner is another key. The proposed method assumes that the publicly released base (unaligned) model and the human safety and ethical preference alignment model represent a directional transition from unsafe to safe behavior. According to the principle of matrix projection, the projection matrix between the unaligned base model and the aligned 
instruction model is computed as the safety projection matrix $W_{\text{safe}}$ as follow:
\begin{equation}
W_{\text{safe}} = \frac{(W_{\text{align}} - W_{\text{base}}) \cdot (W_{\text{align}} - W_{\text{base}})^T}{\text{Dim}(W_{\text{align}} - W_{\text{base}})}
\label{eq:projection_matrix}
\end{equation}

Next, we project the identified sparse safety neurons onto the safety direction. This work employs the LoRA method~\cite{hu2022lora} for efficient fine-tuning on downstream tasks. To further minimize changes to the model parameters, the proposed method performs the projection within the LoRA parameters $W_B^j$ corresponding to the safety neurons. The computation is formulated as follow:
\begin{equation}
W_B^j =\text{Proj}_{\text{safe}}(W_B^j)= \text{Mask}_l \cdot W_{\text{safe}} \cdot W_B^j
\label{eq:masked_projection}
\end{equation}

\subsection{Continual Safety Neuron Projection}
Facing increasingly severe safety challenges, we expect the proposed method to maintain the capability for continuous adaptation to multidimensional safety concerns. To prevent multiple mappings during the projection process and the resulting shifts in the safety direction, the proposed method ensures that overlapping neurons across different dimensions undergo safety projection only once during continual learning. Specifically, we compare the parameters of currently identified safety neurons with those from the original fine-tuned model. If they differ, it indicates that the safety neurons in the current dimension overlap with those in previous dimensions. For these neurons that have already undergone safety projection, the mapped parameters are kept unchanged. Conversely, the newly added safety neuron $\mathcal{N}_{\text{new}}$ parameters in the current dimension are projected onto the safety direction, as formulated in Eq. 8.
\begin{equation}
W_B^j = \mathbb{I}\left( j \in \mathcal{N}_{\text{new}} \right) \cdot \text{Proj}_{\text{safe}}(W_B^j) + \mathbb{I}\left( j \notin \mathcal{N}_{\text{new}} \right) \cdot W_B^j
\label{eq:continual_proj_indicator}
\end{equation}
Where $\mathbb{I}()$ denotes the indicator function. Sparse heterogeneous projections maintain consistency in safety direction, enhancing the safety of new dimensions while preventing catastrophic forgetting of previously adjusted safety dimensions.
\begin{table*}[t]
  \centering
  {\setlength{\tabcolsep}{6pt}
  \begin{tabular}{clccccr}
  \hline\hline
  \textbf{Model} & \textbf{Method} & \textbf{\makecell{Edit\\ Param}} & \textbf{\makecell{GPT-4o \\Judger}} & \textbf{\makecell{Llama3.1-\\405B Judger}} & \textbf{\makecell{Keyword\\ ASR}}& \textbf{\makecell{AlpacaEval\\ Winrate}} \\
  \hline\hline
  \multirow{6}{*}{\makecell{Llama-3-8B\\-Instruct}} 
    & LoraFinetune  & 0.00\% & 2.94 & 3.11 & 55\% & 100.00\% \\
    & SelfReminder & 0.00\% & 2.83 & 2.88 & 47\% & 44.53\% \\
    & GoalPriority & 100.00\% & 2.12 & 2.30 & 35\% & 41.83\% \\
    & SafeLoRA & 10.00\% & 1.55 & 1.51 & 30\% & 47.37\% \\
    & Wanda & 6.78\% & 1.79 & 1.84 & 30\% & 54.15\% \\
    & \textbf{Our MSCP} & \textbf{5.38\%} & \textbf{1.02} & \textbf{1.27} & \textbf{14\%} & \textbf{54.61\%} \\
  \hline
  \multirow{6}{*}{\makecell{Qwen-2.5-7B\\-Instruct}} 
    & LoraFinetune  & 0.00\% & 2.28 & 2.91 & 25\% & 100.00\% \\
    & Self-Reminder & 0.00\% & 2.01 & 2.24 & 27\% & 50.00\% \\
    & GoalPriority & 100.00\% & 2.10 & 2.10 & 30\% & 52.53\% \\
    & SafeLoRA & 11.00\% & 1.95 & 1.93 & 22\% & 50.98\% \\
    & Wanda & 9.42\% & 1.51 & 1.52 & 16\% & 51.40\% \\
    & \textbf{Our MSCP} & \textbf{4.67\%} & \textbf{1.37} & \textbf{1.36} & \textbf{14\%} & \textbf{54.52\%} \\
  \hline\hline
  \end{tabular}}
  \caption{Safety and Utility Across Different Defense Methods on Alpaca-Finetuned Models.}
  \label{tab1}
\end{table*}

\begin{table}[t]
  \centering
  \setlength{\tabcolsep}{1pt}
  \renewcommand{\arraystretch}{1.0}
  \begin{tabular}{lcccc}
    \hline\hline
    \textbf{Method} & \textbf{Judger} & \textbf{Score/ASR} & \textbf{Param} & \textbf{Accuracy}\\
    \midrule
    \multirow{3}{*}{\makecell{Lora\\ Finetune}} 
      & GPT-4o             & 3.95 & \multirow{3}{*}{0\%}    & \multirow{3}{*}{54.20\%} \\
      & Llama3.1-405B      & 3.32 &                         &                           \\
      & Keywords           & 58\%    &                         &                          \\
    \hline
    \multirow{3}{*}{\makecell{Goal \\Priority}} 
      & GPT-4o             & 3.81 & \multirow{3}{*}{100\%}  & \multirow{3}{*}{45\%}    \\
      & Llama3.1-405B      & 3.11 &                         &                           \\
      & Keywords           &  55\%    &                         &                           \\
    \hline
    \multirow{3}{*}{SafeLoRA} 
      & GPT-4o             & 2.18 & \multirow{3}{*}{10.00\%} & \multirow{3}{*}{52.20\%}  \\
      & Llama3.1-405B      & 2.26 &                         &                            \\
      & Keywords           &  57\%    &                         &                           \\
    \hline
    \multirow{3}{*}{Wanda} 
      & GPT-4o             & 3.28 & \multirow{3}{*}{7.63\%} & \multirow{3}{*}{52.60\%}  \\
      & Llama3.1-405B      & 2.91 &                         &                            \\
      & Keywords           &  65\%    &                         &                         \\
    \hline
    \multirow{3}{*}{\textbf{Our MSCP} }
      & GPT-4o             & 1.94 & \multirow{3}{*}{5.46\%} & \multirow{3}{*}{53.20\%} \\
      & Llama3.1-405B      & 1.93 &                         &                            \\
      & Keywords           &  45\%    &                         &                           \\
    \hline\hline
  \end{tabular}
  \caption{Comparison of Safety and Utility on GSM8K-Finetuned Llama-3-8B-Instruct Models.}
  \label{tab2}
\end{table}

\section{Experiments}
\subsection{Experimental Settings}

\textbf{Datasets and Models}. To evaluate the effectiveness, we adopt Llama3.1-8B-Instruct~\cite{meta2024llama3} and Qwen2.5-7B-Instruct~\cite{qwen2024qwen25} using the parameter-efficient fine-tuned LoRA approach on two distinct datasets: the instruction-following semantic Alpaca~\cite{taori2023stanford} and the mathematical reasoning GSM8K~\cite{cobbe2021training}. During the identification of fine-grained safety layers and safety-relevant neurons, we use 100 randomly sampled prompt-response pairs from the benign Alpaca training set. The safety dataset comprises 100 harmful prompts from the “Goal” module of the JailbreakBench~\cite{mazeika2024harmbench}, along with safe responses generated by GPT-4o~\cite{openai2024gpt4o}. For defense evaluation, we assess overall safety performance using 100 randomly sampled general safety prompts from BeaverTails~\cite{ji2023beavertails}. This dataset also enables multi-dimensional safety evaluation to assess the continual alignment capabilities of MSCP. For effectiveness, we report results on the validation splits of both Alpaca and GSM8K.

\textbf{Baseline Methods}. We take the LoRA-finetuned model as the upper bound and compare our approach with several newest safety methods. Among them, self-reminders~\cite{xie2023defending} are a prompt-based defense method through system-mode. Goal Priority~\cite{zhang2023defending} combines discriminator enhancement and fine-tuning by transforming jailbreak prompts into safe refusals through targeted model tuning. SafeLoRA~\cite{hsu2024safe} and Wanda~\cite{wei2024assessing} are both based on the same post-fine-tuning safety defense paradigm as ours, with SafeLoRA editing safety-relevant layers and Wanda targeting safety-relevant neurons.
\begin{table*}[t]
  \centering
  \renewcommand{\arraystretch}{1} 
  \setlength{\tabcolsep}{6pt}     
  \begin{tabular}{ccccccccc}
    \hline\hline
    \diagbox[width=3cm]{\textbf{Safety CL}}{\textbf{MSCP CL}} 
      & \textbf{Judge Method}
      & \textbf{\makecell{Lora\\ Finetune}}
      & \textbf{\makecell{Universal\\ Safety} }
      & \textbf{\makecell{Animal\\Abuse}} 
      & \textbf{\makecell{Child \\Abuse}}
      & \textbf{Terrorism}  \\
    \hline\hline

    \multirow{3}{*}{\textbf{Animal Abuse}}
      & GPT-4o   & 1.64 & 1.32 & 1.32 & 1.20 & 1.18  \\
      & Llama3.1-405B & 1.98 & 1.67 & 1.67 & 1.30 & 1.18  \\
      & Keywords   & 48\% & 28\% & 28\% & 26\% & 8\%   \\
    \hline

    \multirow{3}{*}{\textbf{Child Abuse}} 
      & GPT-4o   & 2.04 & 1.18 & 1.06 & 1.00 & 1.08  \\
      & Llama3.1-405B & 2.08 & 1.44 & 1.26 & 1.10 & 1.26  \\
      & Keywords   & 38\% & 14\% & 14\% & 16\% & 4\%   \\
    \hline

    \multirow{3}{*}{\textbf{\makecell{Controversial\\Politics}}} 
      & GPT-4o   &   1.10   &   1.02   &    1.00  & 1.00     & 1.12  \\
      & Llama3.1-405B &  1.26    & 1.14     &   1.08   &  1.01  & 1.13  \\
      & Keywords   &    72\%  & 56\%     &  56\%    &  56\%    & 72\%  \\
    \hline

    \multirow{3}{*}{\textbf{Self Harm}} 
      & GPT-4o   & 1.56 & 1.58 & 1.20 & 1.14 & 1.13  \\
      & Llama3.1-405B & 1.74 & 1.42 & 1.23 & 1.06 & 1.35  \\
      & Keywords   & 66\% & 84\% & 50\% & 44\% & 20\%  \\
    \hline

    \multirow{3}{*}{\textbf{Terrorism}} 
      & GPT-4o   & 2.36 & 1.74 & 1.18 & 1.08 & 1.10  \\
      & Llama3.1-405B & 2.34 & 2.04 & 1.18 & 1.26 & 1.14  \\
      & Key   & 20\% & 22\% & 8\%  & 4\%  & 4\%   \\
    \hline

    \multirow{3}{*}{\textbf{Privacy}} 
      & GPT-4o   & 2.36 & 1.56 & 1.30 & 1.44 & 1.27  \\
      & Llama3.1-405B & 2.24 & 1.70 & 1.60 & 1.38 & 1.58  \\
      & Keywords   & 34\% & 14\% & 18\% & 24\% & 36\%  \\
    \hline\hline

    \textbf{Utility} & \textbf{Winrate} 
      & \textbf{100\%} & \textbf{54.61\%} & \textbf{50\%} & \textbf{59.12\%} & \textbf{55.87\%}  \\
    \hline\hline
  \end{tabular}
  \caption{Evaluating Safety and Utility under Multi-Dimensional Continual MSCP Projections.}
  \label{tab3}
\end{table*}
\textbf{Evaluation Metrics}. We follow~\cite{chao2024jailbreaking} to evaluate the proposed method from safety and utility. For safety, we adopt both LLM-based safety judger and keyword-based attack success rate (ASR). The LLM judger score ranges from 1 to 10, with higher scores indicating greater unsafe behavior. For utility, we use the AlpacaEval~\cite{alpaca_eval} semantic evaluation benchmark to compute the WinRate between the original LoRA-finetuned model and the MSCP-modified model on the Alpaca dataset. We use accuracy to evaluate performance on the GSM8K dataset. 

\subsection{Performance on Static Safety Concerns}
\textbf{Enhancing safety defenses.} In terms of safety judged by closed-source GPT-4o and open-source Llama3.1-405B, MSCP consistently achieves the lowest harmfulness scores across both Alpaca-finetuned models (1.02 / 1.27 on Llama-3-8B and 1.37 / 1.36 on Qwen-2.5-7B), clearly outperforming alternative methods such as Goal Priority ($\geq$2.10), Self Reminder ($\geq$2.01), and SafeLoRA ($\geq$1.51) as Tab. \ref{tab1}. Moreover, MSCP achieves the lowest keyword ASR on Alpaca-finetuned models (14\% on both Llama and Qwen), demonstrating superior robustness compared to Goal Priority (35\%, 30\%) and Wanda (30\%, 16\%). These results indicate MSCP’s effectiveness in mitigating a broad spectrum of safety risks. Additionally, on the GSM8K-finetuned model, MSCP maintains strong performance, with GPT-4o and Llama3.1-405B harmfulness scores of 1.94 and 1.93, respectively—outperforming all other defense-capable baselines as Tab. \ref{tab2}.

\begin{figure*}[t]
\centering
\includegraphics[width=1.8\columnwidth]{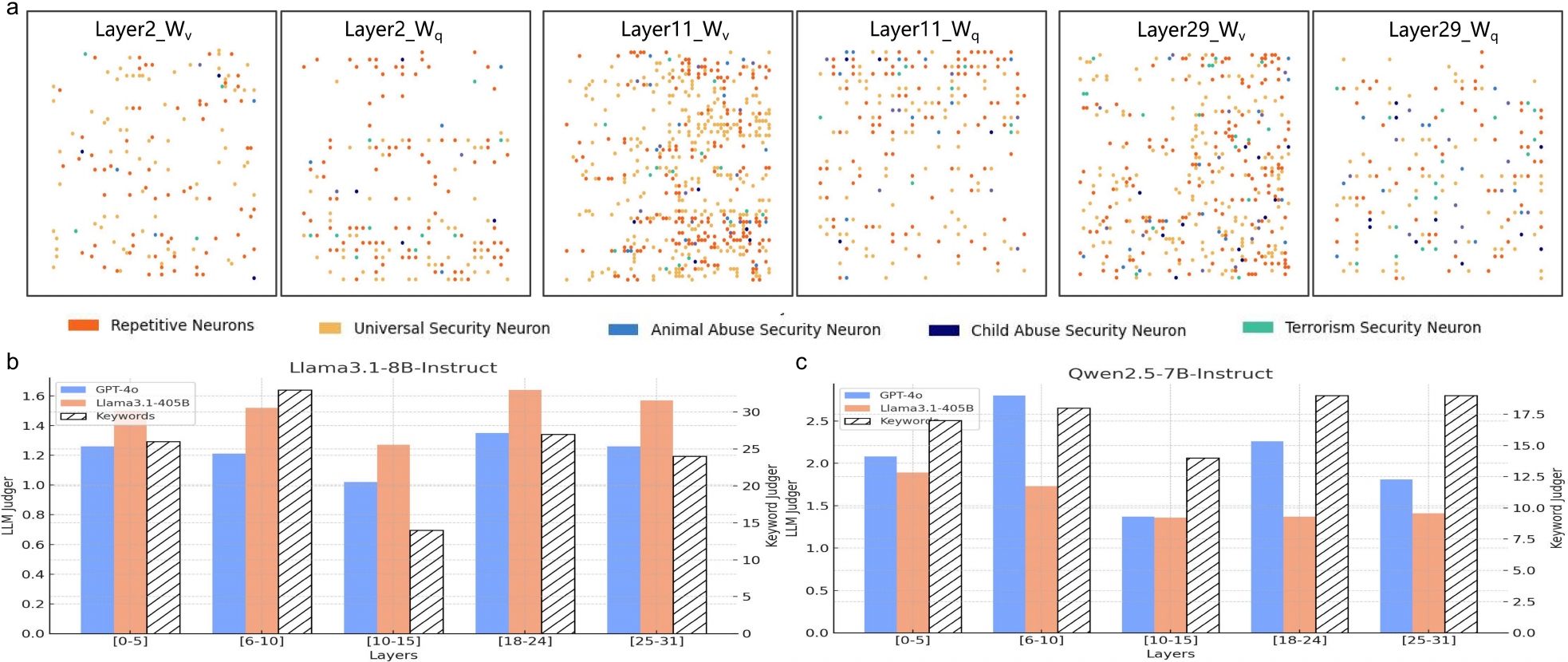} 
\caption{\textbf{a)}:Visualization of safe neuron selection across different safety dimensions.\textbf{b) and c)}: In both LLaMA3.1-8B-Instruct and Qwen2.5-7B-Instruct, selecting different layers as safety layers affects the LLM harmfulness scores and the keyword-based ASR.}
\label{fig3}
\end{figure*}

\textbf{Maintaining utility.} MSCP achieves AlpacaEval Winrates of 54.61\% and 54.52\% on Llama-3-8B and Qwen-2.5-7B fine-tuned models, respectively, surpassing SafeLoRA (47.37\%, 50.98\%) and Wanda (54.15\%, 51.40\%). This indicates that MSCP preserves semantic understanding and response quality without significant degradation. Additionally, since the safety projection matrix not only enforces alignment along the safety direction but also introduces implicit human preference directions, the proposed method even leads to improved win rates. On the GSM8K mathematical reasoning task, MSCP achieves an accuracy of 53.20\%, outperforming SafeLoRA (52.20\%) and Wanda (52.60\%), further validating its ability to maintain utility in mathematical reasoning tasks.

\textbf{Minimal parameter editing.} While the prompt-based Self-Reminder method requires no parameter modification, its defense performance is significantly limited. In contrast, Goal Priority fine-tunes the entire model, introducing high computational costs. Compared to other parameter-editing approaches such as SafeLoRA and Wanda, MSCP makes substantially fewer changes—only (Llama-3-8B: 5.38\% and  Qwen-2.5-7B:4.67\%), resulting in minimal disruption to model utility. Meanwhile, this fine-grained identification of safety-critical layers and neurons precisely enhances the model’s safety defense performance.

\subsection{Performance on Continual Safety Concerns}

\subsubsection{Safety Dimension Generalization.} To address the increasingly complex safety challenges faced by LLMs, we evaluate the proposed Continual MSCP method using multiple risk categories from the BeaverTails safety dataset~\cite{ji2023beavertails}. As shown in Tab.\ref{tab3}, the horizontal axis represents models obtained after continual projections along safety directions targeting different dimensions, while the vertical axis reports harmfulness scores across various safety data. Results demonstrate that after the first projection using only Universal Safety data, the harmfulness scores across all evaluated dimensions decrease significantly(e.g., GPT-4o score drops from 2.04 to 1.18 on Child Abuse, and from 2.36 to 1.56 on Privacy). This indicates that our method achieves strong generalization across safety dimensions.

\subsubsection{Continual reduction in harmfulness.} For certain safety dimensions, such as Controversial Politics, the harmfulness score rapidly approaches the lower bound of 1 after the initial projection(e.g., GPT-4o: 1.10 → 1.02), indicating near-optimal safety performance. In contrast, other dimensions still exhibit room for improvement. Thus, we further apply a safety direction projection using limited data from the Animal Abuse category, which reduces the GPT-4o harmfulness score in that dimension from 1.64 to 1.32.  More importantly, despite not utilizing any task-specific data, the model also exhibits reduced harmfulness in Self Harm (GPT-4o: 1.58 → 1.20), Terrorism (GPT-4o: 1.74 → 1.18), and Privacy (GPT-4o: 1.56 → 1.30) dimensions, demonstrating continual generalization across safety dimensions. Subsequently, we perform continual MSCP projections using data from Child Abuse and Terrorism. The model achieves further safety improvements on these dimensions, with the overall harmfulness score judged by GPT decreasing to an average value of 1.15. Furthermore, it maintains the safety gains on earlier dimensions, confirming that our method supports continual learning without forgetting.

\subsubsection{Continued Preservation of Utility.} Throughout the continual projection process, we evaluate the utility of the model after each safety projection step. Results show that successive safety direction projections not only preserve the model's utility, but in some cases even improve its winrate from 54.61\% after the Universal Safety projection to 55.87\% after the final Terrorism projection, indicating that the responses become more aligned with human preferences.

\subsubsection{Visualization of Safety Neurons across Dimensions.}
Fig. \ref{fig3}a visualizes the distribution of safety neurons selected from Universal Safety data and those identified from various safety dimensions. Orange-red nodes represent safety neurons that overlap. The results show that, for each dimension introduced during continual learning, the majority of selected safety neurons overlap with those from the universal safety projection. Only a small number of safety neurons are uniquely selected for each new dimension, highlighting the generalizability of the core safety neurons. Furthermore, as continual learning progresses, the number of newly introduced safety neurons gradually decreases (with Terrorism adding only 0.73\% of parameter modifications). This indicates that our method continues to improve safety while gradually reducing parameter modifications, leading to a stable model state. Therefore, MSCP does not introduce significant interference with the model's generalizability.

\subsection{Ablation Study}
To investigate the advantages of the proposed multi-Level safety localization, we divide the models into several layer intervals. Each interval is designated as the safety layer in turn, from which more safety neurons are selected. We then evaluate the performance of the modified models on two metrics: LLM harmfulness score and keywords ASR, as Fig. \ref{fig3}.

Results on LLaMA3.1-8B-Instruct model shown in Fig. \ref{fig3}b indicate that replacing the safety layer in shallow layers (layers 1–5 or 5–10) does not significantly reduce the harmfulness scores or the ASR. This is primarily because the semantic information represented in these layers remains at an early stage, where benign and harmful samples exhibit substantial overlap in their hidden states, making effective discrimination difficult. Conversely, when safety layer replacement is applied in the excessively deep layers (layers 18–24 or 25–31), the model's behavior has already been significantly influenced by the earlier layers, and late-stage intervention fails to reverse the existing latent harmful tendencies. Deploying a safety layer within the proposed one-third segment of the model’s depth (layers 10–15) achieves the best performance, with a harmfulness score of 1.02 and an ASR of 14\%. This placement effectively balances semantic representation and behavioral control, enabling more precise localization of safety neurons and thereby enhancing both safety and efficacy. Consistent results were also obtained in the Qwen2.5-7B-Instruct model.

\section{Conclusion}
In this work, we introduce Fine-Grained Safety Neurons with Training-Free Continual Projection (MSCP), a novel defense framework designed to mitigate safety risks introduced by fine-tuning LLMs. By characterizing the coordinated interactions between global and localized safety representations across multiple levels, we precisely identify highly sparse neuron clusters that govern safety-sensitive behaviors, while minimizing interference with general-purpose task performance. Through sparse, training-free projections along safety directions, our approach not only enhances defense effectiveness under static conditions but also supports continual adaptation to emerging safety threats. Extensive experiments across multiple models and benchmarks validate that MSCP consistently improves safety with minimal parameter edits and negligible degradation in utility. However, its ability to continual learning across more complex and diverse safety dimensions remains an area for further investigation. Overall, our work highlights the importance of fine-grained model introspection for robust safety alignment and opens new directions for scalable, maintenance-free safety defenses in the evolving LLM landscape.

\bibliographystyle{named}
\bibliography{ijcai26}

\end{document}